\documentclass[twoside]{article}
\usepackage[accepted]{aistats2017}
\usepackage[compress, round]{natbib}
\usepackage[utf8]{inputenc} 
\usepackage[T1]{fontenc}    
\usepackage{hyperref}       
\usepackage{url}            
\usepackage{booktabs}       
\usepackage{nicefrac}       
\usepackage{microtype}      
\usepackage{resizegather}

\usepackage{amsthm, amsmath, amssymb, amsfonts, bm, graphicx, multirow,  scalerel, slashbox}
\usepackage{algorithm, algorithmic}
\newcommand{\DSZ}{\mathcal{D} \raisebox{0.5mm}{\text{$\underset{\scriptscriptstyle T+1}{\times}$}} (\bm{s}_n\circ\bm{z}_n)}

\theoremstyle{definition}

\newcommand{\ind}{\mathsf{i}}
\newcommand{\smim}[1]{{\small $#1$}}

\begin{document}


\runningauthor{A Stevens, Y Pu, Y Sun, G Spell, L Carin}

\twocolumn[
\aistatstitle{Tensor-Dictionary Learning with Deep Kruskal-Factor Analysis}
\aistatsauthor{ Andrew Stevens \hspace{-1em} \And Yunchen Pu \And \hspace{-1em} Yannan Sun \And \hspace{-2em} Gregory Spell \And \hspace{-1em} Lawrence Carin  }
\aistatsaddress{\hspace{1.5em} PNNL \& Duke U. \hspace{-1em}\And Duke Univ. \And \hspace{-1em} PNNL \And \hspace{-2em} Duke Univ. \And \hspace{-1em} Duke Univ. }
]

\begin{abstract}
A multi-way factor analysis model is introduced for tensor-variate data of any order. Each data item is represented as a (sparse) sum of Kruskal decompositions, a Kruskal-factor analysis (KFA). KFA is nonparametric and can infer both the tensor-rank of each dictionary atom and the number of dictionary atoms. The model is adapted for online learning, which allows dictionary learning on large data sets. After KFA is introduced, the model is extended to a deep convolutional tensor-factor analysis, supervised by a Bayesian SVM. The experiments section demonstrates the improvement of KFA over vectorized approaches (\emph{e.g.}, BPFA), tensor decompositions, and convolutional neural networks (CNN) in multi-way denoising, blind inpainting, and image classification. The improvement in PSNR for the inpainting results over other methods exceeds 1dB in several cases and we achieve state of the art results on Caltech101 image classification.
\end{abstract}

\section{Introduction}
The first data analysis applications of tensors saw the development of the Tucker and canonical decomposition (CANDECOMP) in chemometrics and the parallel factor model (PARAFAC) in linguistics; the CANDECOMP/PARAFAC is also called the Kruskal decomposition (KD). Most recently, tensor methods have been applied in signal/image processing and machine learning \citep{tensor}.

Tensor decompositions are limited in their representation size, and the number of atoms/factor-loadings can be at most the rank of the tensor (\emph{i.e.}, they are not overcomplete). Following on the success of the K-SVD dictionary learning algorithm \citep{ksvd}, several tensor-dictionary learning approaches have been proposed, \emph{e.g.} \citep{sedil}.  The goal of overcomplete dictionary learning is to find a set of atoms that can represent each data item as a weighted sum of a small subset of those atoms. A dictionary is called overcomplete when the number of atoms $K$ is greater than the dimension $P$ of each data item, this is closely related to frame theory \citep{frames}. Overcompleteness is desirable because it offers robustness to noise, increased sparsity, and improved interpretability \citep{overcomplete}.
An overview of non-tensor dictionary learning approaches can be found in \citep{dlrev}.

Previous work in overcomplete tensor-dictionary learning has focused on separable models \citep{sepfilt,sedil,dlsamp}. Imposing separable structure means that each atom is a rank-1 tensor. Moreover, they consider only ``shallow'' representations and are limited to 3-way data.
In this paper we develop Bayesian nonparametric models for tensor-variate data with tensor-variate dictionary atoms and learn the rank of each atom.

The concept of overcompleteness becomes unclear when we move to high-order data. Vectors can be represented as a sum of basis elements, so an overcomplete dictionary gives more elements than needed to represent a vector. In contrast, Matrices (2-way tensors) can be represented as a sum of rank-1 matrices (\emph{i.e.}, SVD). If a particular matrix is $m_1\times m_2$, an overcomplete vectorized representation will have more than $m_1m_2$ atoms, but a ``rank-overcomplete'' representation only needs more than $\min\{m_1,m_2\}$ atoms.

To address the limitations of current (tensor) dictionary learning approaches, we propose a nonparametric tensor-factor analysis model. The proposed Kruskal-factor analysis (KFA) is compatible with tensors of any order and capable of performing blind inpainting. Moreover, we provide extensions to deep convolutional learning and implement a minibatch learning approach. Finally, experimental results on image processing and classification tasks are shown with state of the art results for 2D \& 3D inpainting and Caltech 101, and evidence that rank is an important aspect of overcomplete modeling.

\section{Preliminaries}
\textbf{Dictionary learning: } Formally, dictionary learning is a factorization of a data tensor $\mathcal{X}$. If the data are column vectors we have $\bm{X}=\bm{DW}+\bm{E}$, where the $N$ data vectors are combined into the matrix $\bm{X}\in\mathbb{R}^{P\times N}$, the dictionary $\bm{D}\in\mathbb{R}^{P\times K}$, the weight matrix $\bm{W}\in\mathbb{R}^{K\times N}$, and $\bm{E}$ is the residual/noise. When $K>P$ it is necessary to impose that $\bm{W}$ is sparse to ensure identifiability. 

Beta-process factor analysis (BPFA) is a Bayesian nonparametric model for dictionary learning \citep{BPFA1}.
In BPFA we assume that the noise is Gaussian and impose sparsity using a truncated beta-Bernoulli process. The BPFA model is specified as
\begin{align}
\small
\begin{split}
\bm{x}_n&\sim\mathcal{N}(\bm{Dw}_n, \gamma_\epsilon^{-1}\bm{I}_P),\quad 
\bm{w}_n = \bm{s}_n \circ \bm{z}_n\\
\bm{d}_k &\sim \mathcal{N}(0, P^{-1}\bm{I}_P),\quad
\bm{s}_n\sim\mathcal{N}(0,\gamma_s^{-1}\bm{I}_K)\\
z_{kn}&\sim\mathsf{Bernoulli}(\pi_k),\,\,
\pi_k \sim \mathsf{Beta}\left(a_\pi/K,\, b_\pi(K-1)/K\right)
\end{split}
\raisetag{1.75\baselineskip}
\end{align}
where $\gamma_\epsilon, \gamma_{s}$ have noninformative gamma priors with all hyperparameters set to $10^{-6}$, the prior for $\pi_k$ is nearly uniform with $a_\pi=K$, $b_\pi=1$, and $\circ$ is the Hadamard elementwise product. The probabilities $\pi_k$ determine how likely an atom is to be used and their hyperparameters directly affect sparsity. Other priors for the weights have been proposed in \citep{LevyBPFA,shrinkBPFA, treeBPFA, sigmoidBPFA}. Gaussian process atoms were used in \citep{GPBPFA}.

\textbf{Tensor decomposition: }
A tensor is a mathematical object describing linear transformation rules, and once we fix a basis we can represent a tensor as a multidimensional array of numbers. In data analysis, it is common to call a multidimensional array a tensor. An entry of a  $T$-way or order-$T$ tensor $\mathcal{X}\in\mathbb{R}^{m_1\times \cdots \times m_T}$ is indicated by a position vector $\ind = [\ind_1,\ldots,\ind_T]$, where $\ind_t$ is an integer on $[1,m_t]$; we denote the $\ind^{\text{th}}$ entry of the tensor $\mathcal{X}$ as $x_\ind$. We use $\otimes$ to represent the Kronecker product. We also use the mapping $\bm{x} = \text{vec}(\mathcal{X})$, defined by $x_{\ind}=\text{vec}(\mathcal{X})_{\ind_1 + \ind_2m_1 + \cdots + \ind_Tm_1\cdots m_{T-1}}$.

If we consider a matrix $\bm{X}\in\mathbb{R}^{m_1\times m_2}$ we can write it as \smim{ \bm{X} = \sum_{r=1}^{m_1\wedge m_2} \lambda_r \bm{u}_{r}^{(1)} \bm{u}_{r}^{(2)\top} }, where \smim{\lambda_r\in\mathbb{R},\,\bm{u}_{r}^{(1)}\in\mathbb{R}^{m_1},\,\bm{u}_{r}^{(2)}\in\mathbb{R}^{m_2}}. This is the SVD. A generalization of the SVD to a $T$-way tensor $\mathcal{X}$ is called the canonical polyadic decomposition (PD). The canonical PD is written as
\smim{\mathcal{X} = \sum_{r=1}^{R_M} \lambda_r \bigotimes_{t=1}^T \bm{u}_r^{(t)}}, where $R_M$ is the maximum rank of tensors in \smim{\mathbb{R}^{m_1\times\cdots\times m_T}$, $\bm{u}_{r}^{(t)}\in\mathbb{R}^{m_t}}. The rank of a tensor is the number of nonzero $\lambda_r$. 

Recently, the multiplicative gamma process CANDECOMP/PARAFAC (MGP-CP) \citep{MGPCP}, a Bayesian nonparametric KD, was proposed. MGP-CP can additionally infer the rank of a tensor. In MGP-CP, the MGP \citep{Dunson} imposes that the singular values should shrink in absolute value as $r$ increases. The MGP shrinks the variance of a zero mean Gaussian (the prior for $\lambda_r$), so that for large $r$ the singular value will be near zero with high probability. When a singular value shrinks enough (\emph{e.g.}, $\lambda_r<10^{-6}$) we assume it is zero. The numerical rank is given by the number of nonzero $\lambda_r$ after thresholding. The MGP-CP is described entrywise, \smim{ x_{\ind} = \sum_{r=1}^R \lambda_r \prod_{t=1}^T u_{\ind_t r}^{(t)} + \epsilon_{\ind}}
is the $\ind^{\text{th}}$ entry of $\mathcal{X}$, $R$ is the maximum desired rank and $\epsilon_{\ind}$ is Gaussian noise. The MGP-CP is specified as
\begin{align}
\small
\begin{split}
x_{\ind} &\sim \mathcal{N}\left(\scaleobj{0.7}{\sum_{r=1}^R} \lambda_r \scaleobj{0.7}{\prod_{t=1}^T} u_{\ind_t r}^{(t)},\, \tau_\epsilon^{-1}\right),\, \bm{u}_{r}^{(t)} \sim \mathcal{N}(0, \omega_{rt}^{-1}\bm{I}_{m_t})\\
 \lambda_r &\sim \mathcal{N}(0, \tau_{kr}^{-1}),\quad \tau_{kr} = \scaleobj{0.7}{\prod_{i=1}^r} \delta_i,\quad \delta_i \sim \mathsf{Gamma}(\alpha, 1)
\end{split}
\end{align}
and $\tau_\epsilon, \omega_{rt}$ have noninformative gamma priors with all hyperparameters set to $10^{-6}$. The shape parameter $\alpha$ for the MGP is usually set to 2 or 3.  If $\alpha > 1$, the error in \smim{\sum_{r=1}^R \lambda_r \prod_{t=1}^T u_{\ind_t r}^{(t)}} converges to zero exponentially fast as $R\to\infty$ \citep{MGPCP}.

\textbf{Tensor Rank: }
The maximum rank $R_M$ is usually not $\min_t m_t$ like the matrix case.
Finding the canonical PD of a tensor is a difficult computational problem. It is possible to find PD with a preset $R$ (that could be different from the actual rank), these are called KD.

A useful characterization of tensors is balance, which we use to set $R$ for KFA. Assuming $m_T$ is the largest dimension, a tensor is balanced when \smim{m_T \le \prod_{t=1}^{T-1}m_t - \sum_{t=1}^{T-1}(m_t-1)} and unbalanced when \smim{m_T \ge 1+\prod_{t=1}^{T-1}m_t - \sum_{t=1}^{T-1}(m_t-1)}. An unbalanced tensor has generic rank \smim{R_G = \min\{ m_T,  \prod_{t=1}^{T-1} m_t\} \le R_M}, similar to matrices. Balanced tensors are expected to have generic rank \smim{R_E\!=\!\left\lceil \prod_t m_t \big/ \left(\sum_t m_t -T+1\right)\right\rceil \!\le\! R_M} \citep{tensorRank}.

With the notions of tensor rank introduced we can define rank-overcomplete.
A tensor factorization is rank-overcomplete when the number of (tensor) atoms is greater than the rank of the tensor.
We note that this definition also applies to slices of tensors, \emph{e.g.} data items that have been stacked into a larger tensor. For example, in 2D \smim{ \bm{X} = \sum_{k=1}^{K} \lambda_k \bm{u}_{k}^{(1)} \bm{u}_{k}^{(2)\top} }, is an overcomplete decomposition if \smim{K>\min\{m_1, m_2\}}.

\section{Kruskal Factor Analysis}

\begin{figure*}
\centerline{\includegraphics[width=0.95\textwidth]{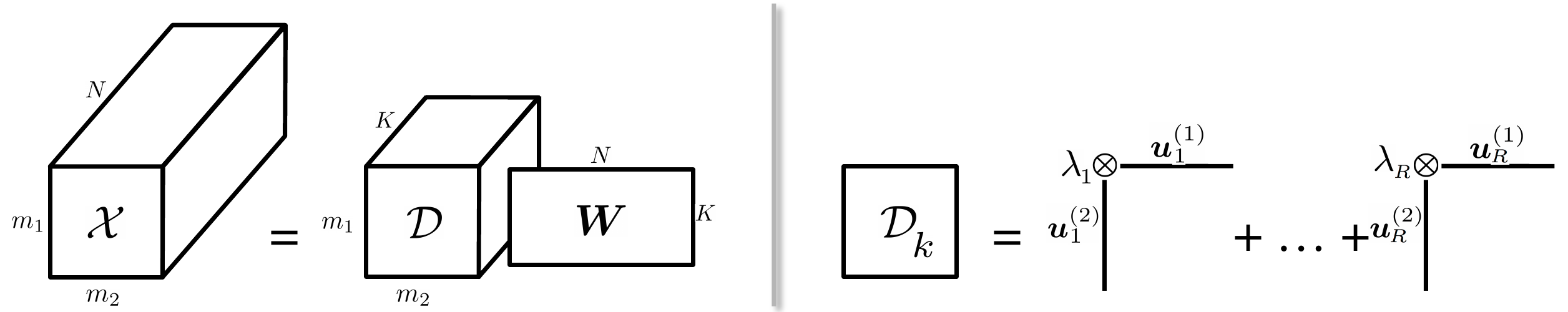}}
\caption{\small An illustration of the KFA Model for 2-way data. The left equation shows the mode-$3$ tensor product generalization of dictionary learning. The right equation shows the representation of dictionary atom $k$ using the KD, for 2-way data this is the SVD. The illustration clearly shows the reduction in the number of parameters compared to a non-decomposed dictionary.
Each atom in a non-decomposed dictionary will have \smim{\prod m_t} parameters, while our approach has \smim{R(1 + \sum m_t)}. The ability to specify $R$ gives our model greater flexibility. Moreover, by applying shrinkage on $\lambda_r$ and learning the shrinkage rate $\alpha$, our model automatically adjusts model complexity during learning.}
\label{fig:model}
\vskip -1.5em
\end{figure*}
The most natural generalization of dictionary learning to tensors is to model the data as the product of a dictionary tensor and a weight matrix along the $(T+1)^\text{st}$ dimension of the dictionary. We define the $\text{mode-}t$ tensor product $\times_t$ with an example: the $\text{mode-}3$ product of a tensor $\mathcal{D}\in\mathbb{R}^{m_1\times m_2\times K}$ and a matrix $\bm{W}\in\mathbb{R}^{K\times N}$ yields $\mathcal{X}=\mathcal{D}\times_3 \bm{W}\in\mathbb{R}^{m_1\times m_2 \times N}$ with \smim{x_{\ind_1\ind_2\ind_3} = \sum_{k=1}^K d_{\ind_1\ind_2 k}w_{k\ind_3}}. The factorization of $\mathcal{X}$ is illustrated for 2-way data in Figure~\ref{fig:model}.

We extend dictionary learning to arbitrary order tensors by inferring the ``singular'' vectors/values of each atom via KD. The Kruskal representation for each atom is used in the likelihood. It is important to note that when we learn the model we are not replacing the dictionary with a low-rank approximation. The singular vectors/values are drawn from their posteriors, thus they are directly affected by the data; if the data requires full-rank atoms they will be learned. We call this new model Kruskal-factor analysis (KFA). The model is specified as follows: 
\begin{align}
\small
\begin{split}
\mathcal{X}_n &\sim\, \mathcal{N}( \text{\small vec}[\mathcal{D}\raisebox{0.5mm}{\text{$\underset{\scriptscriptstyle T+1}{\times}$}} (\bm{s}_n\circ\bm{z}_n)], \gamma_\epsilon^{-1} \bm{I}),\,
d_{\mathsf{i}k} = \scaleobj{0.7}{\sum_{r=1}^R} \lambda_{rk} \scaleobj{0.7}{\prod_{t=1}^T} u_{\mathsf{i}_t r}^{(kt)}\\
u^{(kt)}_{\mathsf{i}_t r} &\sim \mathcal{N}(0, m_t^{-1}),\,\,
\lambda_{kr} \sim \mathcal{N}(0,\tau_{kr}^{-1}),\,\, \tau_{kr} = \scaleobj{0.7}{\prod_{i=1}^r} \delta_{ki}\\
s_{kn} &\sim\mathcal{N}(0,\gamma_s^{-1}), \quad z_{kn}\sim\mathsf{Bernoulli}(\pi_k)\\
\pi_k &\sim \mathsf{Beta}\left(a_\pi/K,\, b_\pi(K-1)/K\right),\quad
\delta_{ki} \sim \mathsf{Gam}(\alpha,1) 
\end{split}
\raisetag{2.25\baselineskip}
\end{align}
where $\mathcal{X}_n\in\mathbb{R}^{m_1\times \cdots \times m_T}$ is the $n^\text{th}$ data item and $d_{\ind k}$ is the $\ind  = [\ind_1,\ldots,\ind_T]$ element of the $k^\text{th}$ atom. The hyperparameters for $\gamma_\epsilon,\gamma_s$ are set to $10^{-6}$ to give a noninformative prior and $a_\pi = K, b_\pi=1$. The shape parameter $\alpha$ for singular value shrinkage is inferred, this is discussed below. Note that the negative log-likelihood is simply \smim{\|\mathcal{X} - \mathcal{D} \!\text{\raisebox{2pt}{$\underset{\scriptscriptstyle T+1}{\times}$}}\!(\bm{S}\circ\bm{Z})\|_F^2}, the atom decompositions impose tensor structure and the priors correspond to regularizers. By design, our model is fully locally conjugate and can be implemented efficiently as a Gibbs sampler. A Gibbs sampler produces a Markov chain whose stationary distribution is the joint distribution of the model, and the mode of a set of samples (from the chain) is a maximum a posteriori (MAP) solution to the corresponding regularized optimization problem.

In order to gain intuition about the model, consider the case when we fix $R=1$. Every atom $\mathcal{D}_k$ will be rank~1, so each data item is a sum of rank~1 tensors. This is a KD where the rank~1 tensors are selected from a dictionary. The rank of the output tensor is determined by the number of nonzero entries in $\bm{z}_{n}$ ($\min \{\text{nnz}(\bm{z}_n), R_M\}$), with $\bm{s}_n$ acting as singular values. Increasing $R$ yields: \smim{\mathcal{X}_n = \sum_{k,r} s_{nk}z_{nk}\lambda_{rk} \bigotimes_t \bm{u}_r^{(kt)}  }, which is a KD using rescaled singular values and  $s_{nk}z_{nk}$.

\vspace{0.5em}
\textbf{Dictionary updates: }
With a careful derivation we have found an update procedure that seamlessly converts BPFA into a tensor model. The naive approach, specifying the likelihood elementwise, as in MGP-CP, is poorly suited to computation, since the singular values/vectors must be updated for every entry of the data tensor---making the dictionary update complexity \smim{O\left(NKT^2R(1+\sum m_t)\prod m_t \right)} for a single Gibbs iteration (without including the expensive residual computation). In our implementation, entire atoms are updated at once. Moreover, the atom-wise updates clearly show the overhead of the KD. The first step in the update is to compute the mean and covariance just as in BPFA, these are then used to sample all of the atom parameters. With the following reparameterizations 
\begin{gather}
\begin{aligned}[c]
\scaleobj{0.9}{d_{\mathsf{i}k}} &=  \scaleobj{0.9}{\lambda_{rk}\scaleobj{0.8}{\prod_{t'\neq t}} u_{\mathsf{i}_{t'} r}^{(kt')}  u_{\mathsf{i}_t r}^{(kt)}+  \scaleobj{0.8}{\sum_{r'\neq r}} \lambda_{r'k}\scaleobj{0.8}{\prod_{t=1}^T} u_{\mathsf{i}_t r'}^{(kt)}  }
\!\!\!\!\!&=&\,a_{\mathsf{i}_t} u_{\mathsf{i}_t r}^{(kt)} + b_{\mathsf{i}_t}\\
&= \scaleobj{0.8}{\prod_{t=1}^T} u_{\mathsf{i}_t r}^{(kt)}  \lambda_{rk} + \scaleobj{0.8}{\sum_{r'\neq r}} \lambda_{r'k} \scaleobj{0.8}{\prod_{t=1}^T} u_{\mathsf{i}_t r'}^{(kt)}
&=&\,f_{\mathsf{i}}\lambda_{rk} + g_{\mathsf{i}},\quad
\end{aligned}
\raisetag{2.25\baselineskip}
\end{gather}
we derive closed form updates for dictionary atoms:
\begin{gather}
\begin{aligned}[c]
u^{(kt)}_{\mathsf{i}_t r} &\sim \mathcal{N}(\hat{\mu}, \hat{\omega}^{-1})\\
\hat{\omega} &= m_t + \scaleobj{0.8}{\sum_{\mathsf{i}:\mathsf{i}_t=m}} \gamma^{\mathcal{D}_k}_{\mathsf{i}_t k}\left( a_{\mathsf{i}_t} \right)^2\\
\hat{\mu} &= \frac{1}{\hat{\omega}} \scaleobj{0.8}{\sum_{\mathsf{i}:\mathsf{i}_t=m}} \left(\mu^{\mathcal{D}_k}_{\mathsf{i}_t k} - b_{\mathsf{i}_t} \right) a_{\mathsf{i}_t}
\end{aligned}\,\,
\begin{aligned}[c]
\lambda_{kr} &\sim \mathcal{N}(\hat{\nu},\hat{\tau}^{-1})\\
\hat{\tau} &= \tau_{kr} + \gamma_\epsilon \scaleobj{0.8}{\sum_{n,\mathsf{i}}} \left( f_{\mathsf{i}} s_{kn} \right)^2 \\
\hat{\nu} &= \frac{\gamma_\epsilon}{\hat{\tau}}\scaleobj{0.8}{\sum_{n,\mathsf{i}}} \left( h_{\mathsf{i}n} - g_{\mathsf{i}}s_{kn} \right)f_{\mathsf{i}} s_{kn},\quad
\end{aligned}
\raisetag{2.75\baselineskip}
\end{gather}
where \smim{\gamma^{\mathcal{D}_k}_{\mathsf{i}_t k}, \mu^{\mathcal{D}_k}_{\mathsf{i}_t k}} are the BPFA updates for the precision and mean (without multiplying the covaraince) and \smim{\mathcal{H} = \mathcal{X} - \mathcal{D} \!\text{\raisebox{2pt}{$\underset{\scriptscriptstyle T+1}{\times}$}}\!(\bm{S}\circ\bm{Z})} is the current residual.

The complexity for sampling $\mathcal{D}$ in BPFA is \smim{O(KM^2 \sum n_k^2)} (with inpainting \citep{BPFA3}), where \smim{M=\prod m_t} and $n_k$ is the number of data items using atom $k$. This arises from the residual updates (which includes an elementwise matrix product and an outer product of matrices), which dominate. For a $T$-way tensor KFA adds \smim{O(KT^2R(1+\sum m_t))} operations, so the complexity for sampling $\mathcal{D}_k$ in KFA is the same as BPFA (since \smim{M^2 \sum n_k^2 \gg T^2R(1+\sum m_t)}, for practical N). The weight matrices $\bm{S}$ and $\bm{Z}$ have the same updates in BPFA and KFA, so the complexity of one Gibbs iteration of KFA is the same as BPFA.

\vspace{0.5em}
\textbf{Learning the shape parameter $\alpha$: }
The shape parameter is critical to model quality because it determines the shrinkage rate for the singular values. We found that fixing $\alpha$ produces suboptimal results. Updating $\alpha$ while the model is learned is important because it adjusts the complexity of the dictionary. Note that a similar approach was taken in \citep{MGPCP}, where the rank of the tensor was increased or decreased \textit{ad hoc} depending on whether the singular values are larger than or smaller than preset thresholds.

KFA specifies a prior for $\alpha$, which connects the dictionary atoms. The conjugate prior for a gamma distribution with known rate $\beta=1$ is given by $p(\alpha;a,b) \propto \frac{a^{\alpha -1}}{\Gamma(\alpha)^b}$, so the posterior distribution is
\begin{align}
\begin{split}
p(\alpha|\{\delta_{ki}\},a,b) &= \mathsf{Gamma}(\delta_{ki};\alpha,1) p(\alpha;a,b)\\
&\propto \left(a\raisebox{0.2mm}{\scaleobj{0.9}{\prod_{k,i}}}\delta_{ki}\right)^{\alpha-1}\Gamma(\alpha)^{-b-RK}.
\end{split}
\label{eqn:alpha} \end{align}
Although this prior gives straightforward updates, it is not a standard distribution to sample from. We use numerical integration to perform inverse-CDF sampling. Using the conditional MAP solution for $\alpha$ also works well and is simpler. The MAP solution is closed form: \smim{\psi^{-1}\left( \frac{\log(a)+\sum_{k,i}\delta_{ki}}{b+RK} \right)}, where $\psi^{-1}$ is the inverse digamma function.

We choose the hyperparameters $a=10^6,\, b=10^{-6}$ and set the initial value for $\alpha=10^6$. This causes the model to gradually increase from rank~1 atoms up to the inferred minimal rank. Without sampling $\alpha$, the parameters $R$ and $\alpha$ must be selected manually. In our approach we set $R=R_E-1$ or $R=R_G-1$, depending on whether each data item is balanced or not, and let the model learn $\alpha$. We reduce the rank by 1 to prevent overfitting. 

\vspace{0.5em}
\textbf{Scalability: }
We use the Bayesian conditional density filter (BCDF) \citep{bcdf} to allow our model to scale to large datasets. Essentially, BCDF allows MCMC models to be trained online or by iterating over subsets of training data. This application of BCDF is not the same as the original authors intended. In their work, only one epoch was used.
Our BCDF approach removes the need to load an entire dataset into memory. The details are given in the supplement.

\section{Deep Convolutional KFA}
Instead of matrix-vector multiplication between the atoms and the sparse weights we can use convolution. This will imbue the model with translational invariance; specializing BPFA with a convolutional likelihood is called convolutional factor analysis (CFA) \citep{CFA}. In the setting of images we use the 2D (spatial) convolution, and for multi-channel images the convolution is applied separately to each channel. The likelihood for CFA is
\smim{ \bm{X}_n \sim\mathcal{N}\left(\sum_k\bm{D}_k * \bm{W}_{kn}, \gamma_\epsilon^{-1}\bm{I}_P\right)},
which can be converted into a tensor model by applying KD to $\bm{D}_k$ and applying the same priors as KFA. Note that the weights $\bm{W}_{kn}$ are matrices and the data  $\bm{X}_n\in\mathbb{R}^{M_x \times M_y\times M_c}$ and dictionary atoms $\bm{D}_k\in\mathbb{R}^{m_x\times m_y\times M_c}$ are tensors ($e.g.$, $M_c=3$ for RGB images). Usually, the dimensions of the dictionary atoms are much smaller than the image dimensions ($m_x \ll M_x$ and $m_y\ll M_y$). The spatially-dependent activation weights for dictionary atom $k$, image $n$, are $\bm{W}_{kn}\in\mathbb{R}^{(M_x-m_x +1)\times(M_y - m_y +1)}$. More details can be found in \cite{CFA,pu1,DGDN}. The dictionary atoms in CFA, although they are tensors, are sampled from a multivariate Gaussian. 

In a deep architecture, the set of weights $\{\bm{W}_{kn}\}_{k=1}^K$ for image $n$ are represented in terms of (different) convolutional dictionary atoms.
When an $L$-layer deep model is built, the input of the $\ell^{th}$ layer is usually composed of a pooled version of the output of the layer below, the $(\ell-1)^{st}$ layer. We can formulate this deep deconvolutional model via these two contiguous layers:
\begin{gather}
\small
\begin{aligned}[c]
\bm{X}_n^{(\ell-1)}=\scaleobj{0.8}{\sum_{k=1}^{K_{\ell-1}}} \bm{D}_k^{(\ell-1)} * \bm{W}_{kn}^{(\ell-1)},\,\,
\bm{X}_n^{(\ell)}={\scaleobj{0.8}{\sum_{k=1}^{K_{\ell}}} \bm{D}_k^{(\ell)} * \bm{W}_{kn}^{(\ell)}},\,\,\qquad
\end{aligned}
\raisetag{1.75\baselineskip}
\label{eq:deep}
\\
\begin{aligned}[c]
\bm{W}_{kn}^{(\ell-1)} := {\mathsf {unpool}}(\bm{X}_{kn}^{(\ell)}).
\end{aligned}
\label{eq:unpool}
\end{gather}
The weight matrices \smim{\bm{W}_{kn}^{(\ell)}} in the $\ell^{\text{th}}$-layer become the inputs to the $(\ell+1)^{\text{st}}$-layer after pooling. The input tensor $\bm{X}_{n}^{(\ell+1)}$ is constituted by stacking the $K_{\ell}$ spatially aligned \smim{\bm{X}_{kn}^{(\ell+1)}}. The \smim{(\ell+1)^{\text{st}}}-layer inputs are tensors with the third-dimension of size $K_\ell$, the dictionary size in the $\ell^{\text{th}}$-layer.
This is deep CFA.
When a stochastic unpooling process is employed in (\ref{eq:unpool}), and appropriate priors are imposed on dictionary and feature parameters, the model developed in (\ref{eq:deep}) constitutes a generative model for images called the deep generative deconvolutional network (DGDN)~\citep{DGDN}. 

Within each layer of the deep CFA and DGDN model we employ KFA, resulting in deep CKFA and Kruskal DGDN. A major difference from multiplicative KFA is that both the weights and the dictionary are tensors. Kruskal DGDN can be supervised by connecting a Bayesian SVM \citep{BSVM} to the top layer weights $\bm{W}_{kn}^L$ \citep{DGDN}.

\section{Related Work}
KFA is different from K-SVD and BPFA, since their atoms are vectors, and thus atom-rank is not defined. Moreover, in the experiments we found that BPFA prefers full-rank atoms (determined by first reshaping), while KFA prefers reduced-rank atoms. In addition, the posterior distribution for a vectorized atom in KFA will have a Kronecker structured covariance (with off-diagonal interactions), whereas BPFA atoms have a diagonal posterior covariance.

The main difference of KFA and any Tucker-like decomposition, \emph{e.g.} sparse Bayesian Tucker decomposition (sBTD) \citep{sbtd}, TenSR \citep{tensr}, Tensor Analyzers \citep{tensorAnalyzer}, or MGP-CP, is that KFA is rank-overcomplete
, but Tucker-like decompositions can have at most $R_M$ (a similar condition holds for multilinear-rank). Another difference from decompositions is that KFA infers several tensors simultaneously (the atoms). Further, we learn the rank and sparsity simultaneously. We also extend MGP-CP by learning the shrinkage rate. Without inferring $\alpha$, KFA performance suffers. MGP-CP is very sensitive to this parameter---techniques are proposed in \citep{MGPCP} for tuning $\alpha$. Moreover, the rank being inferred is different (atoms vs. data tensor), and the atoms could be balanced or unbalanced, but for $N\ge M$ the data tensor is usually unbalanced.

Some recent work has been done using a separable ($R=1$) tensor structure for the dictionary. One approach, separable dictionary learning \citep{sedil}, considered two-way data (grayscale image patches) and claimed generalizing to higher order data was straightforward, but did not show any results in this direction. Their optimization algorithm is significantly more complicated than the separable convolution approach in \citep{sepfilt} and for grayscale denoising the performance is not as good as BPFA. The separable multiplicative structure is also considered in \citep{dlsamp}, where the sample complexity for various dictionary learning approaches is considered.

A separate line of work is concerned with high-order cumulant decomposition, which is used in, \emph{e.g.} independant component analysis \citep{cumulant}. Cumulants are symmetric, so all of the factor matrices are identical \citep{symTensor}. This is very different from direct analysis of tensor data.

\section{Experiments}
In this section we show that maintaining the tensor structure of the data is beneficial in several tasks. We compare against vectorized factor analysis to show that keeping structure can boost performance. All experiments use the default parameters discussed above and none of the parameters were tuned or optimized. We implemented our models in Matlab. All latent parameters are initialized randomly, except $\alpha$ which is set to $10^6$ to initialize an essentially separable model. In the tables, bold represents the best result.

\begin{figure*}[!t]
\vskip 0.2in
\begin{center}
{\includegraphics[height=0.37\textwidth]{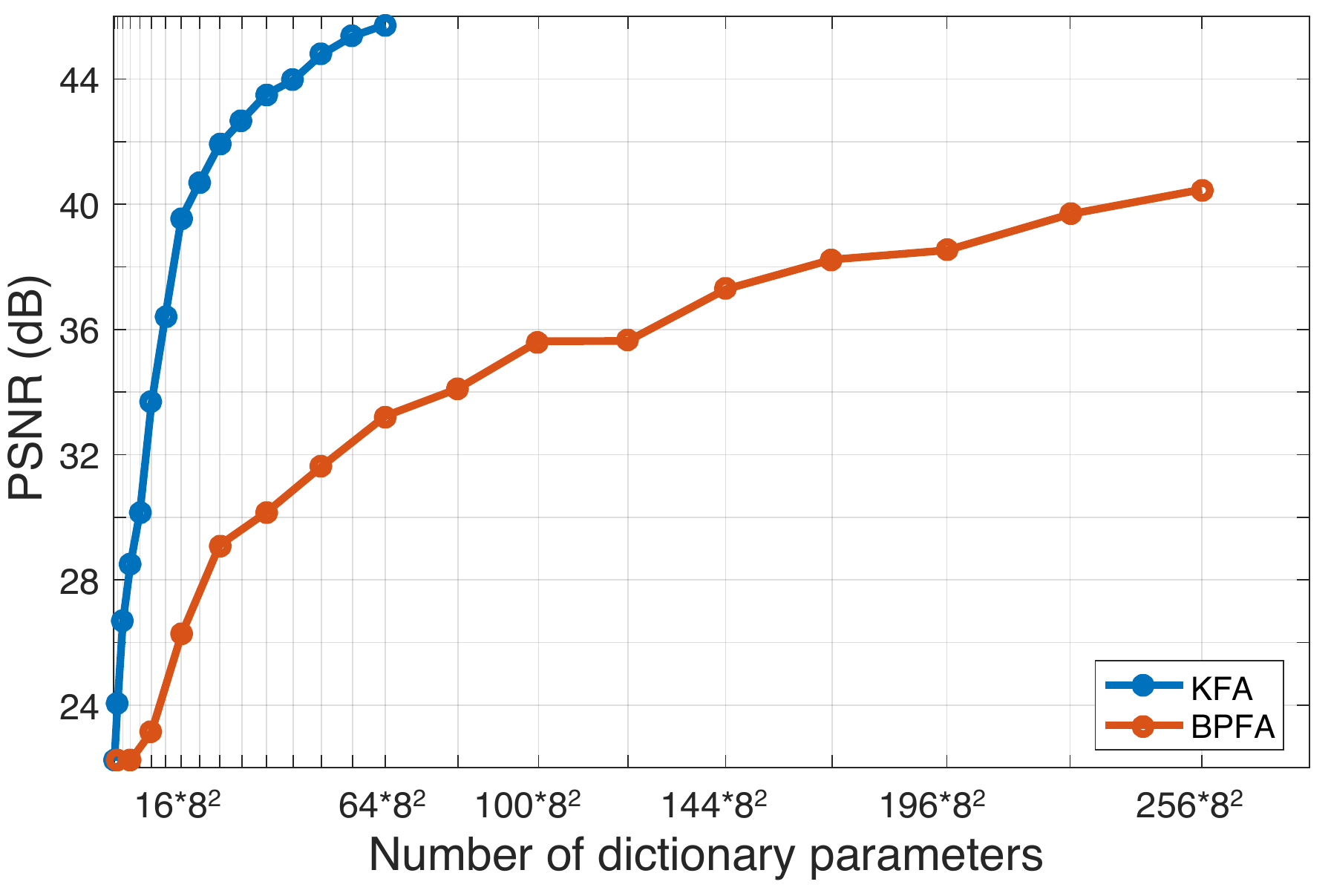}}$\qquad$
{\includegraphics[height=0.37\textwidth]{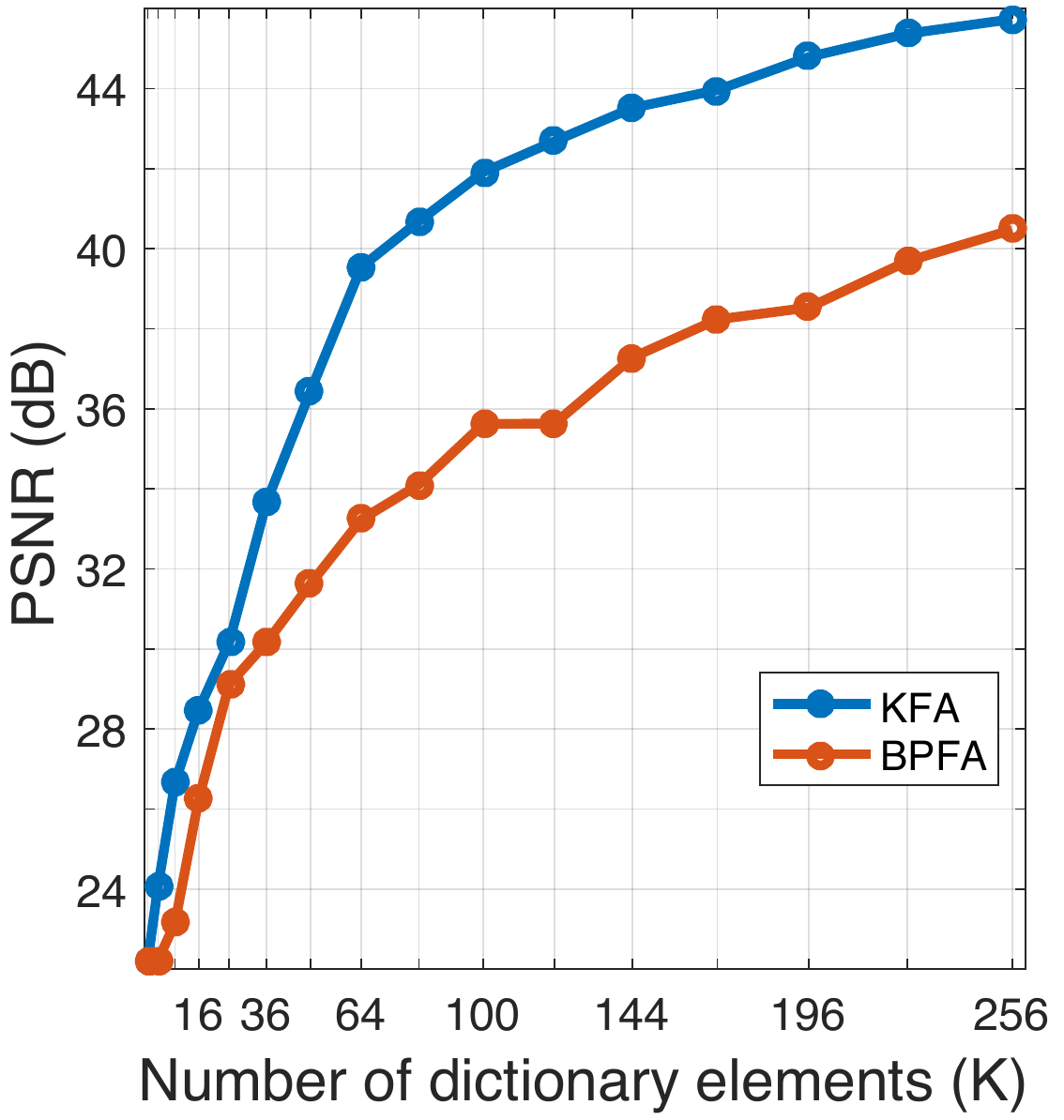}}
\caption{\small Reconstruction quality for varying model complexity. The same data is shown with different $x$-axis scalings: reconstruction PSNR as a function of the number of parameters (left) and as a function of the number of atoms (right). }
\label{fig:complexity}
\end{center}
\vskip -2.0em
\end{figure*}

\subsection{Image denoising and inpainting}
Image denoising is a standard image processing task, and BPFA has been shown to produce very good denoising results without \textit{a priori} knowledge of the noise level. Inpainting is another image processing task, and is a special case of compressive sensing. The goal in the inpainting task is to impute missing pixels. Simultaneous inpainting and denoising has become an important step in scientific imaging \citep{me, Kovarik}. The model details for inpainting and compressive sensing are given in \citep{BPFA3}. Finally, while we do not give direct comparisons to K-SVD \cite{ksvd}, BPFA has been shown to be, in general, superior to K-SVD for denoising.

All of the image processing methods are patch-based. The dataset $\mathcal{X}\in \mathbb{R}^{B\times B\times C \times N}$ is extracted from the image on a regular grid where the corner of each patch is separated from the other patches by $(\Delta,\Delta)$ pixels ($B$ is patch width/height, $C$ is the number of colors). In these experiments 150 samples are obtained, with the first 40 used as burn-in. The remaining samples are reconstructed as an image, and then averaged to give the final reconstruction. This reconstruction approach is a type of Bayesian model averaging, we do not compute a MAP or mean-field solution. For the our stochastic learning approach, termed CDF-KFA, we use 10 epochs with 5 for burn-in, a batch size of 5000, and we update all of the local parameters one time after the first epoch. The time for a single epoch is about 3~times the iteration time for KFA on the full dataset (CDF-KFA has 3$\times$ more updates). For inpainting, the sensing mask pixels are drawn $iid$ from a Bernoulli distribution and the same mask is used for the different algorithms. We measure reconstruction error using peak signal to noise ratio (PSNR). 

\textbf{Grayscale: } The grayscale experiments use a $256\times 256$ Barbara image. The image was chosen so that a direct comparison can be made with published BPFA results. The patch size is $8\times 8$ in these experiments.

We first examine the performance of KFA by varying the complexity (via dictionary size $K$) and robustness to small sample sizes. The results of the complexity experiment are shown in Figure~\ref{fig:complexity}; the dictionary size ranges from 1 to 256, $\Delta=1$, and for KFA $R=1$. We see that KFA, with much fewer parameters, outperforms BPFA. The number of parameters for the KFA dictionary is \smim{KR(1+\sum m_t)=17K} and BPFA has \smim{K\prod{m_t}=64K}. KFA also has a sizable jump in performance just above $K=R_M=8$, but before $K=\prod m_t$, indicating that rank-overcompleteness is beneficial. For robustness to small sample sizes we train on datasets with $\Delta = 1,2,4,8$, corresponding to 100\%, 25\%, 6.25\%, and 1.56\% of the patches; results are shown in Table~\ref{tab:datasize} each entry is the mean of 10 runs; the standard deviation for KFA and BPFA was~0.1, for MGP-CP~1.2 and for CDF-KFA~0.6.
We also compare to MGP-CP (using patches) with $R=R_G=64$ and our new update for $\alpha$ using the same number of samples as KFA. KFA performs substantially better than BPFA and KFA is more consistent for smaller datasets. We also find that the MGP-CP, performs better than BPFA.

\begin{table}[ht]
\begin{center}
\vskip -1.0em
\caption{\small Reconstruction PSNR (dB) for fractions of data.}
\vskip 0.2em
\label{tab:datasize}
\begin{small}
\begin{sc}
  \begin{tabular}{l| c | c | c | c }
    $\Delta$&1  & 2 &4  & 8  \\ \hline
    {\small MGP-CP}&       47.23 &         45.68 &         44.82 &         42.76  \\
    {\small BPFA}&         42.48 &         41.43 &         40.21 &         36.88  \\
    {\small CDF-KFA}&      44.56 &         41.14 &         40.83 &         39.01  \\ 
    {\small KFA}&  \textbf{50.36}& \textbf{49.71}& \textbf{47.84}& \textbf{44.26}  
  \end{tabular}
\end{sc}
\end{small}
\end{center}
\vskip -1.0em 
\end{table}

\begin{figure*}[!t]
\begin{center}
\centerline{
\includegraphics[width=0.23\textwidth]{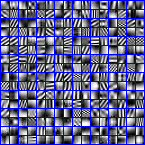}$\,$
\includegraphics[width=0.23\textwidth]{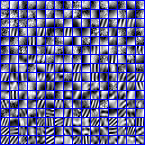}$\,$
\includegraphics[width=0.25\textwidth]{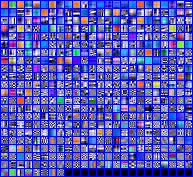}$\,$
\includegraphics[width=0.25\textwidth]{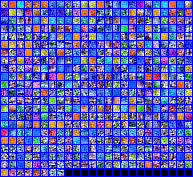}
}
\vskip -0.5em
\caption{\small Dictionaries from clean images. Left to right: Barbara KFA, BPFA; Castle KFA, BPFA. Zoom for detail.}
\label{fig:barbara}
\end{center}
\vskip -3em
\end{figure*}

\begin{figure*}[!t]
\vskip 0.2in
\begin{center}
\centerline{
\includegraphics[width=0.24\textwidth]{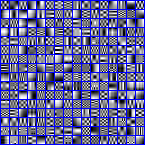}$\,$
\includegraphics[width=0.24\textwidth]{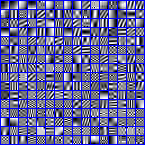}$\,$
\includegraphics[width=0.24\textwidth]{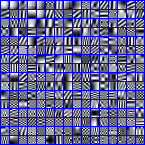}$\,$
\includegraphics[width=0.24\textwidth]{barbara_dict.png}$\,$
}
\vskip -0.5em
\caption{\small Barbara KFA dictionaries: from left to right $R=1,2,4,8$. For $R=1$ KFA learns a very DCT-like dictionary. As the rank increases diagonal structures are added. For $R=8$, most atoms have rank less than~5, and none are full-rank.}
\label{fig:barbaraRank}
\end{center}
\vskip -2.0em
\end{figure*}

The results for simultaneous denoising and inpainting of the Barbara image are listed in Table~\ref{tab:barbara} and example reconstructions for 20\% inpainting are in the SM. We find that KFA generally outperforms BPFA. The BPFA results were reported in \citep{BPFA3}.
\begin{table}[!hb]
\vskip -2.5em
\begin{center}
\caption{\small Reconstruction PSNR (dB) for the Barbara image reduced pixel collection and additive noise. Top: BPFA, Bottom: KFA, Subtable: CDF-KFA.}
\vskip -0.5em
\label{tab:barbara}
\begin{small}
\begin{sc}
  \begin{tabular}{l| c | c | c | c | c }
    \backslashbox{\raisebox{0pt}[\height][0pt]{$\sigma$}}{\hspace{-2em}\vspace{0.5em}{\tiny Pixels}}& 10\%& 20\%& 30\%& 50\%& 100\%  \\ \hline
    \multirow{2}{*}{0}&        23.47&           26.87&          29.83&          35.60&          42.94\\
    &                  \textbf{24.24}&  \textbf{27.92}& \textbf{30.90}& \textbf{36.03}& \textbf{50.43}\\\hline

    \multirow{2}{*}{5}&        23.34&           26.73&          29.27&  \textbf{33.61}&         37.70\\
    &                  \textbf{24.11}&  \textbf{27.43}& \textbf{30.08}&         33.51&  \textbf{38.22}\\\hline

    \multirow{2}{*}{10}&       23.16&           26.07&          28.17&          31.17&          34.31\\
    &                  \textbf{23.51}&  \textbf{27.18}& \textbf{28.54}& \textbf{31.24}& \textbf{34.44}\\\hline
    
    \multirow{2}{*}{15}&       22.66&           25.17&          26.82&          29.31&          32.14\\
    &                  \textbf{23.09}&  \textbf{25.43}& \textbf{27.92}& \textbf{29.36}& \textbf{32.18}\\\hline
    
    \multirow{2}{*}{20}&       22.17&           24.27&          25.62&          27.90&          30.55\\
    &                  \textbf{22.47} & \textbf{24.57}& \textbf{26.09}& \textbf{28.72}& \textbf{30.71}\\\hline
    
    \multirow{2}{*}{25}&       21.68&           23.49&          24.72&          26.79&          29.30\\
    &                  \textbf{21.91} & \textbf{23.96}& \textbf{25.09}& \textbf{27.02}& \textbf{30.17}
\bigskip \\[-0.5em]
    Pixels& 10\%&20\%&30\%&50\%&100\%\\
    $\sigma$& 0&5&10&15&20\\\hline
    &   23.51& 26.80& 27.92& 28.72& 30.17
  \end{tabular}
\end{sc}
\end{small}
\end{center}
\vskip -1em
\end{table}
The dictionaries learned by each algorithm from the uncorrupted image are shown in Figure~\ref{fig:barbara}.
Qualitatively, the KFA dictionary has a larger variety of structures.  
The rank (via SVD, with a threshold of $10^{-6}$) of the KFA dictionary atoms is between 2~and~7 with~96\% having rank~4 or~5. In contrast, every BPFA atom has full rank.
KFA dictionaries for $R=1,2,4,8$ learned from the uncorrupted image are shown in Figure~\ref{fig:barbaraRank}. The dictionaries clearly show that KFA enforces different rank; in higher-rank dictionaries KFA learns both low- and high-rank atoms.

The results for only inpainting are substantially better than BPFA. KFA gains more than 1~dB in 3 of the cases. For denoising cases KFA also produces superior results. In combined denoising and inpainting KFA performs better than BPFA in all but one case.

\textbf{Color: } In the color image experiment we use the castle image. The patch size is $7\times 7\times 3$, $\Delta = 1$, and $K=512$. Denoising and inpainting were performed separately and the reconstruction PSNRs for the castle image are shown in Table~\ref{tab:castle}. The inpainting results improve upon BPFA (reported by \citep{BPFA3}), and the denoising results are comparable. 
We also obtain a PSNR of 50.94 dB on the uncorrupted image using non-overlapping patches. The dictionaries learned by BPFA and KFA are shown in Figure~\ref{fig:barbara}. 
\begin{table}[h]
\begin{center}
\vskip -.5em
\caption{\small Reconstruction PSNR (dB) for the castle image.}
\vskip 2mm 
\label{tab:castle}
\begin{small}
\begin{sc}
  \begin{tabular}{l| c | c | c | c }
    Pixel\%& 20 & 30 & 50 & 80  \\ \hline
    BPFA&         29.12 &         32.02 &         36.45 &         41.51 \\
    CDF-KFA&      28.95 &         31.87 &         37.18 &         44.60 \\
    KFA&  \textbf{29.59}& \textbf{32.46}& \textbf{38.40}& \textbf{49.08} \medskip\bigskip\\ 

    $\sigma$& 5 & 10 & 15 & 25\\ \hline
    BPFA&         40.37 & \textbf{36.24}& \textbf{33.98}& \textbf{31.19}\\
    CDF-KFA&      39.94 &         35.63 &         33.52 &         30.60\\
    KFA&  \textbf{40.58}&         36.20 &         33.73 &         30.54\\ 
  \end{tabular}
\end{sc}
\end{small}
\end{center}
\vskip -2em
\end{table}

\textbf{3-way data: }
Here we show the superior inpainting performance of KFA on two 3-way data sets. The first is the amino acids fluorescence data ($5\times 201 \times 61$) \citep{amino}, with $5\times 50\times 4$ patches, $\Delta=1$, $K=256$, and $R=R_G=20$. We compare to BPFA, MGP-CP (no patches), and sBTD \citep{sbtd}. The second is the brainweb, a 3D MRI image \citep{brain1}, where we use $8\times 8\times 8$ patches, $\Delta=4$, $K=512$, and $R=R_E=24$. For the brainweb data, we compare to BPFA, and sBTD. We also show comparable performance to BM4D for denoising \citep{bm4d}. The results are shown in Tables~\ref{tab:amino}--\ref{tab:mri}. We report PSNR and relative residual square error (RRSE): \smim{\|\hat{\mathcal{X}} - \mathcal{X} \|_F/\|\mathcal{X} \|_F}.

After reviewing the results for grayscale and color image inpainting, the substantial performance gain by KFA over the other tensor methods is not surprising and is due primarily to the fact that we are performing tensor dictionary learning, not tensor decomposition--- in fact, BPFA is comparable to tensor decomposition. For 20\%~pixels, KFA obtains a PSNR of 27.92, 29.58, and 30.46~dB, for gray, color, and MRI data. The reason inpainting performance increases as the dimension grows is that the number of sensed pixels in a patch increases with dimension (about 13 pixels for $8\times 8$, and 102 for $8\times 8\times 8$). We further note that for the amino experiment, KFA is rank-overcomplete, but not overcomplete (\smim{R_G<K<\prod m_t}). 

\begin{table}[!h]
\begin{center}
\vskip -0.5em
\caption{\small Reconstruction quality for amino 10\% pixels.}
\label{tab:amino}
\vskip 2mm 
\begin{small}
\begin{sc}
  \begin{tabular}{l| c | c |c |c}
        &sBTD& MGP-CP& BPFA &KFA\\ \hline
    RRSE &0.026& 0.02542& 0.02752 &\textbf{0.01099}\\
    PSNR &45.40& 45.60& 44.91 & \textbf{52.88}
  \end{tabular}
\end{sc}
\end{small}
\end{center}
\vskip -0.5em
\end{table}

\begin{table}[!h]
\begin{center}
\vskip -1em
\caption{\small Reconstruction PSNR (dB) for 3D MRI.}
\label{tab:mri}
\vskip 2mm 
\begin{small}
\begin{sc}
  \begin{tabular}{l| c | c | c | c}
                   & sBTD&           BM4D&  BPFA& KFA\\ \hline
    $\sigma = 19\%$&  ---& \textbf{29.70}& 29.26& 29.67\\
    $20\%$ pixels&  22.33&            ---& 30.21& \textbf{30.46}
  \end{tabular}
\end{sc}
\end{small}
\end{center}
\vskip -1.00em
\end{table}

\subsection{Deep learning}
In the next two experiments we test deep CKFA and Kruskal DGDN. The rank for these experiments was set to $R=1$ to minimize training time. Even with such a strong restriction the proposed tensor models outperform the non-tensor models. The hyperparameters for our deep models are set as in \citep{pu_nips}; no tuning or optimization was performed. All of our deep models are rank-overcomplete (each layer).

\textbf{Caltech 101: } 
We resize the images to $128\times 128$, followed by local contrast normalization~\cite{Jarrett09ICCV}.
The network in this example has 3~layers.
The dictionary sizes for each layer are set to $K_1=64$, $K_2=125$ and $K_3=128$, and the dictionary atom sizes are set to $16\times 16$, $9\times 9$ and $5\times 5$. 
The size of the pooling regions are $4\times 4$ (layer 1 to layer 2) and $2\times 2$ (layer 2 to layer 3).
For classification, we follow the setup in~\cite{yang09CVPR}, selecting 15 and 30 images per category for training, and up to 50 images per category for testing. The training for the CFA and CKFA models is unsupervised, and the top layer features are used to train an SVM after training the deep model.

\begin{table}[ht]
\begin{center}
\vskip -0.5em
\caption{\small Caltech 101 classification accuracy (\%).}
\label{tab:caltech}
\begin{small}
\begin{sc}
	\begin{tabular}{l| c|c }
		Method& 15 images & 30 images \\
		\hline
		HBP-CFA layer-1 & 53.6 & 62.5 \\
		CKFA layer-1 & 55.7 & 63.2 \\\hline
		Deep CFA  & 43.24 & 53.57 \\
		HBP-CFA layer-1+2 & 58.0 & 65.7 \\
		CKFA layer-2 & 57.4 & 64.5 \\
		CKFA layer-3 & 61.2 & 69.9 \\\hline
        DGDN         & 75.4   & 87.8\\
        Kruskal DGDN  & \textbf{76.1}    & \textbf{88.3}

	\end{tabular}
\end{sc}
\end{small}
\end{center}
\vskip -0.5em
\end{table}

We compare to the hierarchical beta process CFA (HBP-CFA) \citep{CFA}, the pretrained DGDN (\textit{i.e.}, deep CFA), and the DGDN \citep{pu_nips}. The DGDN models have 3-layers. Other deep CNN models outperform our results in this experiment. However, we are simply showing that using a tensor structured dictionary can improve performance. The results for Caltech 101 are shown in Table~\ref{tab:caltech}. The one layer CKFA significantly outperforms Deep CFA and is also better than the single layer HBP-CFA. We also find that the 2-layer deep CKFA achieves comparable performance to the deep HBP-CFA using only the top-layer features. The 2-layer HBP-CFA uses both layers of features in the SVM. We also train a 3-layer CKFA and see that the deeper model continues to extract more discriminative features as layers are added. Kruskal DGDN provides state of the art accuracy among models \emph{not} pretrained on ImageNet \citep{imagenet}.

\begin{figure}[t]
\vskip 2em
\begin{center}
{\includegraphics[height=0.32\textwidth, trim=0 0 0 40]{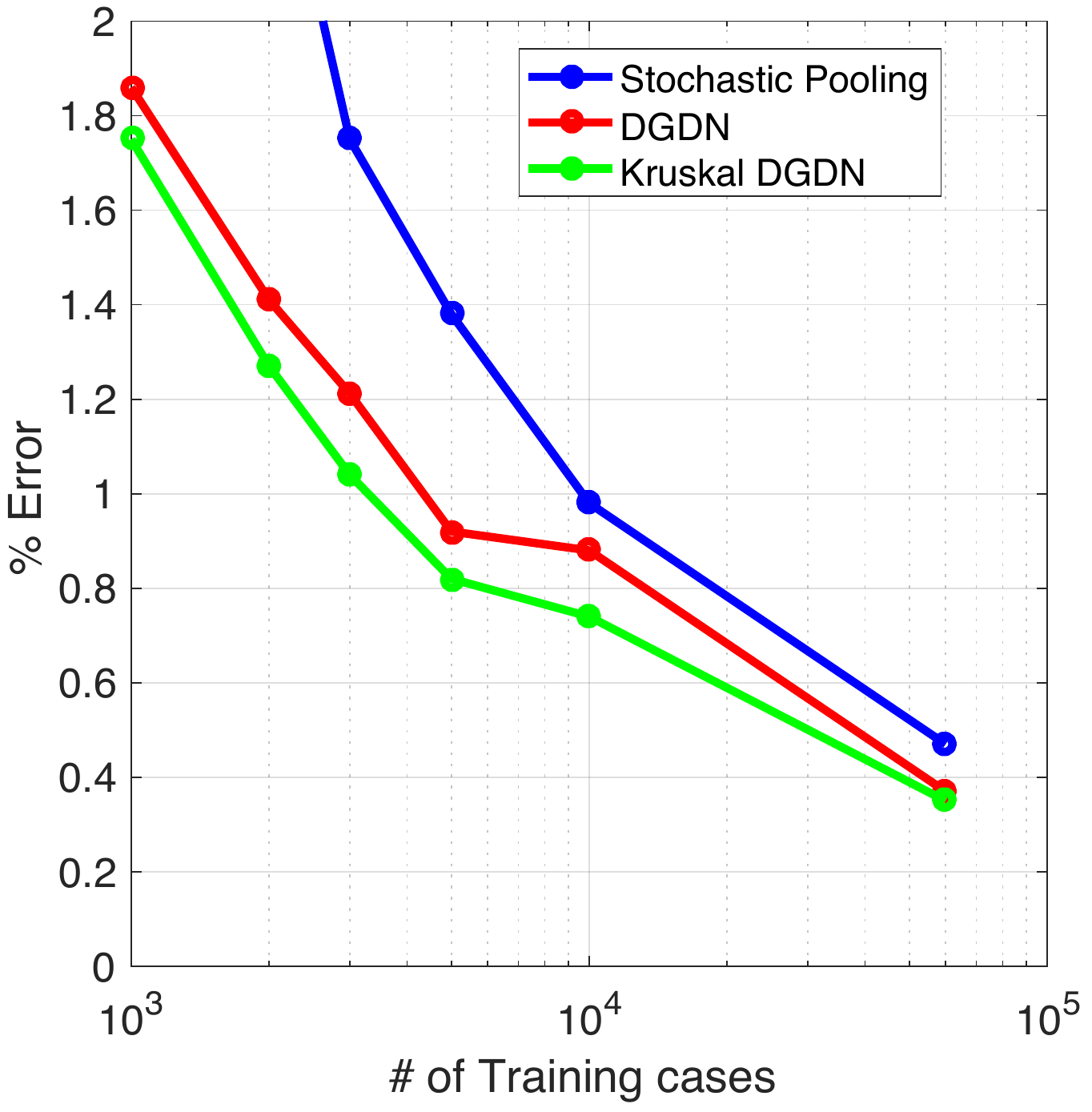}}
\caption{\small MNIST classification error varying training size.}
\label{fig:mnistSamp}
\end{center}
\vskip -1em
\end{figure}

\textbf{MNIST: }
A two layer model is used with dictionary size $8\times 8$ in the first layer and $6\times 6$ in the second layer; the pooling size is $3\times 3$ and the number of dictionary atoms in layers 1 and 2 are $K_1=36$ and $K_2=152$. These numbers of dictionary atoms are obtained by setting the initial number of atoms to a large value ($K_1=50,\,K_2=200$), then removing low-use atoms. 

The results for MNIST are shown in Table~\ref{tab:mnist}. The state of the art MCDNN is shown as a reference point, MCDNN uses a committee of 35 convnets, elastic distortions, and width normalization. In contrast, we train a single 2-layer model using the original training data. The separable Kruskal DGDN outperforms DGDN.   
We also test the robustness of Kruskal DGDN to small data sets. The results are shown in Figure~\ref{fig:mnistSamp}. We compare to the first stochastic pooling work \citep{zeiler} and DGDN. Kruskal DGDN is consistently better than DGDN.

\begin{table}[ht]
\begin{center}
\vskip -1em
\caption{\small MNIST classification error on test-set.}
\vskip 2mm 
\label{tab:mnist}
\begin{small}
\begin{sc}
  \begin{tabular}{ l | c }
  Method & Error \%\\ \hline
  35-CNN \citep{ciresan} & 0.23\\
  6-layer CNN \citep{ciresan2} & 0.35\\
  4-layer CNN \citep{dnnShape}& 0.37\\
  2-layer DGDN \citep{pu_nips} & 0.37\\
  2-layer Kruskal DGDN & 0.35
  \end{tabular}
\end{sc}
\end{small}
\end{center}
\vskip -1.5em
\end{table}

\section{Conclusion}
This paper presented a doubly nonparametric tensor-variate factor analysis model. To the best of our knowledge, this is the first factor analysis model capable of preserving the structure of arbitrary order tensor data. KFA is fully locally conjugate, by design, and can be implemented as a Gibbs sampler.
The concept of rank-overcompleteness was also introduced. We showed that rank has a role in the interaction between sparsity and overcompleteness, but the interplay of these three properties requires further study.
We also explored several extensions of KFA: online learning with BCDF, convolutional factor analysis, deep CFA, and supervised deep CFA using the Bayesian SVM.
The performance of KFA and Kruskal DGDN is promising and has improved state of the art in inpainting and on the Caltech 101 image classification benchmark.

\textbf{Acknowledgments: }
Supported by the Chemical Imaging, Signature Discovery, and Analytics in Motion Initiatives at PNNL and by ARO, DARPA, DOE, NGA, ONR and NSF. PNNL is operated by Battelle Memorial Inst. for the US DOE; contract DE-AC05-76RL01830.

\newpage
\small{
\bibliographystyle{abbrvnat}
\setlength{\bibsep}{0.9pt}
\bibliography{kfa}
}
\newpage
\onecolumn
\section*{Supplementary Material}
\subsection*{Gibbs updates}
KFA Likelihood and priors:
\begin{align}
\begin{split}
\mathcal{X}_n \sim&\, \mathcal{N}( \DSZ, \gamma_\epsilon^{-1} \bm{I}),\,\,
d_{\mathsf{i}k} = \scaleobj{0.7}{\sum_{r=1}^R} \lambda_{rk} \scaleobj{0.7}{\prod_{t=1}^T} u_{\mathsf{i}_t r}^{(kt)}\\
u^{(kt)}_{\mathsf{i}_t r} &\sim \mathcal{N}(0, m_t^{-1}),\,\,
\lambda_{kr} \sim \mathcal{N}(0,\tau_{kr}^{-1}),\,\, \tau_{kr} = \scaleobj{0.7}{\prod_{i=1}^r} \delta_{ki}\\
s_{kn} &\sim\mathcal{N}(0,\gamma_s^{-1}), \quad z_{kn}\sim\mathsf{Bernoulli}(\pi_k)\\
\pi_k &\sim \mathsf{Beta}\left(\nicefrac{a_\pi}{K},\, \nicefrac{b_\pi(K-1)}{K}\right),\quad
\delta_{ki} \sim \mathsf{Gam}(\alpha,1) 
\end{split}
\end{align}
where $\mathcal{X}_n\in\mathbb{R}^{m_1\times \cdots \times m_T}$ is the $n^\text{th}$ data item, and $d_{\ind k}$ is the $\ind  = [\ind_1,\ldots,\ind_T]$ element of the $k^\text{th}$ atom. The precisions $\gamma_\epsilon,\gamma_s$ have gamma priors.

\begin{gather}
\begin{aligned}[c]
\scaleobj{0.9}{d_{\mathsf{i}k}} &=  \scaleobj{0.9}{\lambda_{rk}\scaleobj{0.8}{\prod_{t'\neq t}} u_{\mathsf{i}_{t'} r}^{(kt')}  u_{\mathsf{i}_t r}^{(kt)}+  \scaleobj{0.8}{\sum_{r'\neq r}} \lambda_{r'k}\scaleobj{0.8}{\prod_{t=1}^T} u_{\mathsf{i}_t r'}^{(kt)}  }
\!\!\!\!\!&=&\,a_{\mathsf{i}_t} u_{\mathsf{i}_t r}^{(kt)} + b_{\mathsf{i}_t}\\
&= \scaleobj{0.8}{\prod_{t=1}^T} u_{\mathsf{i}_t r}^{(kt)}  \lambda_{rk} + \scaleobj{0.8}{\sum_{r'\neq r}} \lambda_{r'k} \scaleobj{0.8}{\prod_{t=1}^T} u_{\mathsf{i}_t r'}^{(kt)}
&=&\,f_{\mathsf{i}}\lambda_{rk} + g_{\mathsf{i}}
\end{aligned}
\raisetag{2.25\baselineskip}
\end{gather}

\begin{gather}
\begin{aligned}[c]
u^{(kt)}_{\mathsf{i}_t r} &\sim \mathcal{N}(\hat{\mu}, \hat{\omega}^{-1})\\
\hat{\omega} &= m_t + \scaleobj{0.8}{\sum_{\mathsf{i}:\mathsf{i}_t=m}} \gamma^{\mathcal{D}_k}_{\mathsf{i}_t k}\left( a_{\mathsf{i}_t} \right)^2\\
\hat{\mu} &= \frac{1}{\hat{\omega}} \scaleobj{0.8}{\sum_{\mathsf{i}:\mathsf{i}_t=m}} \left(\mu^{\mathcal{D}_k}_{\mathsf{i}_t k} - b_{\mathsf{i}_t} \right) a_{\mathsf{i}_t}\\
\gamma^{\mathcal{D}_k} &= \gamma_{\epsilon}\bm{I}_P\sum_{n=1}^N s_{kn}^2z_{kn}^2 \\
\mu^{\mathcal{D}_k} &= \gamma_{\epsilon}\sum_{n=1}^N s_{kn}z_{kn}\tilde{\bm{x}}_n^{\setminus k}\\
\tilde{\bm{x}}_n^{\setminus k}  &= \text{vec}\left[  \mathcal{X}_n - \DSZ + s_{kn}z_{kn}\mathcal{D}_k     \right]
\end{aligned}
\raisetag{2.75\baselineskip}
\end{gather}

\begin{gather}
\begin{aligned}[c]
\lambda_{kr} &\sim \mathcal{N}(\hat{\nu},\hat{\tau}^{-1})\\
\hat{\tau} &= \tau_{kr} + \gamma_\epsilon \scaleobj{0.8}{\sum_{n,\mathsf{i}}} \left( f_{\mathsf{i}} s_{kn} \right)^2 \\
\hat{\nu} &= \frac{\gamma_\epsilon}{\hat{\tau}}\scaleobj{0.8}{\sum_{n,\mathsf{i}}} \left( h_{\mathsf{i}n} - g_{\mathsf{i}}s_{kn} \right)f_{\mathsf{i}} s_{kn}\\
h_{\mathsf{i}n}  &= \mathcal{X}_{\mathsf{i}n}  - [\DSZ]_{\mathsf{i}}
\end{aligned}
\raisetag{2.75\baselineskip}
\end{gather}

\begin{gather}
\begin{aligned}[c]
p(s_{ki}|-) &\sim \mathcal{N}(\mu_{s_{ki}}, \Sigma_{s_{ki}})\\
\Sigma_{s_{ki}} &= (\gamma_s + \gamma_{\epsilon}z_{kn}^2\bm{d}_k^T\bm{d}_k)^{-1}\\
\mu_{s_{ki}}  &= \gamma_{\epsilon}z_{kn}\bm{d}_k^T \tilde{\bm{x}}_n^{\setminus k}\\
\bm{d}_k &= \text{vec}[\mathcal{D}_k]
\end{aligned}
\raisetag{2.75\baselineskip}
\end{gather}

\begin{gather}
\begin{aligned}[c]
z_{ki} &\sim \text{Bern}\left(\frac{p_1}{p_0 + p_1}\right)\\
p_1 &= \pi_k\exp[-\frac{\gamma_{\epsilon}}{2}(s_{kn}^2\bm{d}_k^T\bm{d}_k -s_{kn}\bm{d}_k^T \tilde{\bm{x}}_n^{\setminus k}) ]\\
p_0 &= 1 - \pi_k
\end{aligned}
\raisetag{2.75\baselineskip}
\end{gather}

\begin{gather}
p(\gamma_{\epsilon}|-) \sim \Gamma\left( c + \frac{PN}{2},\, d + \frac{1}{2}\sum_{n=1}^N \|\mathcal{X}_n - \DSZ)\|_F^2\right)
\end{gather}

\begin{gather}
p(\gamma_{s}|-) \sim \Gamma\left( e + \frac{KN}{2},\, f + \frac{1}{2}\sum_{n=1}^N \bm{s}_n^T\bm{s}_n\right)
\end{gather}

\begin{gather}
p(\pi_k|-) \sim \text{Beta}\left(\frac{a}{K} + \sum_{n=1}^N z_{kn},\, b\frac{K-1}{K} + N - \sum_{n=1}^N z_{kn}\right)
\end{gather}

\subsection*{Bayesian Conditional Density Filter for KFA}
We record surrogate conditional sufficient statistics (SCSS) for the global parameters that directly interact with the data or local parameters ${\mathcal{D}, \gamma_\epsilon, \gamma_s, \pi_k}$. The SCSS are given by the posterior updates of the global parameters. The SCSS are accumulated for each subset of data, and then applied in the subsequent update as the prior parameters. We reset the SCSS after the number of data items seen is the same as the size of the dataset. In our implementation we update the model parameters 3~times using a single subset of data. The SCSS are held fixed until the 3${}^\text{rd}$ iteration. This allows stale local parameters $\{\bm{S},\bm{Z}\}$ to be refreshed before updating the SCSS. The algorithm below describes the process.

\vskip 2em
\begin{algorithmic}
  \FOR{$epoch=1$ \TO $numEpoch$}
    \STATE Randomly partition $\mathcal{X}$ into minibatches $\mathcal{X}^{(i)}$ of size $batchSize$
    \STATE Set SCSS to prior parameters
    \FORALL{$\mathcal{X}^{(i)}$}
      \STATE Update $\bm{S}^{(i)},\bm{Z}^{(i)}$ and all hyperparameters with KFA
      \STATE Do a standard KFA update twice (including the dictionary) 
      \STATE Update SCSS ($\forall k$):
      \STATE $\qquad  \bm{\mu}^{\mathcal{D}_k}_{\text{SCSS}} =  \bm{\mu}^{\mathcal{D}_k}_{\text{SCSS}} + \bm{\mu}^{\mathcal{D}_k (i)}  $ 
      \STATE $\qquad  \bm{\gamma}^{\mathcal{D}_k}_{\text{SCSS}} =  \bm{\gamma}^{\mathcal{D}_k}_{\text{SCSS}} + \bm{\gamma}^{\mathcal{D}_k (i)}  $ 
      \STATE $\qquad \pi_{ka\text{SCSS}} = \pi_{ka\text{SCSS}} + \sum_n z_{kn}^{(i)}$
      \STATE $\qquad \pi_{kb\text{SCSS}} = \pi_{kb\text{SCSS}} + N-\sum_n z_{kn}^{(i)}$
      \STATE $\qquad  \gamma_{sa{\text{SCSS}}} = \gamma_{sa{\text{SCSS}}} + KN^{(i)}/2$ 
      \STATE $\qquad  \gamma_{sb{\text{SCSS}}} = \gamma_{sb{\text{SCSS}}} + \frac{1}{2}\sum_n \bm{s}_n^{(i)\top} \bm{s}_n^{(i)}$ 
      \STATE $\qquad  \gamma_{\epsilon a{\text{SCSS}}} = \gamma_{\epsilon a{\text{SCSS}}} + PN^{(i)}/2$ 
      \STATE $\qquad  \gamma_{\epsilon b{\text{SCSS}}} = \gamma_{\epsilon b{\text{SCSS}}} + \frac{1}{2}\sum_n \|\mathcal{X}_n^{(i)} - \mathcal{D} \raisebox{0.5mm}{\text{$\underset{\scriptscriptstyle T+1}{\times}$}} (\bm{s}_n^{(i)}\circ\bm{z}_n^{(i)}) \|_F^2 $
    \ENDFOR
    \STATE Reset SCSS to original prior parameters
    \IF{$epoch == 1$} \STATE Update $\bm{S},\bm{Z}$ and all hyperparameters using the full data set (note that every column is independent, so this can be done in parallel). \ENDIF
  \ENDFOR
\end{algorithmic}

\newpage
\subsection*{Barbara Images}
\begin{figure}[!h]
\vskip 0.2in
\begin{center}
\includegraphics[width=0.27\textheight]{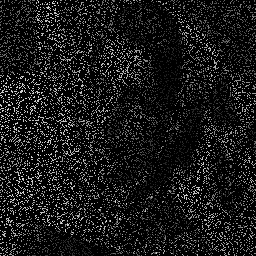}\\
\includegraphics[width=0.27\textheight]{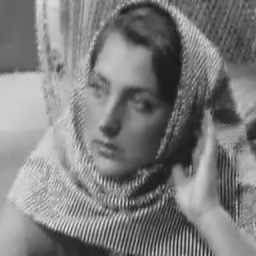}\\
\includegraphics[width=0.27\textheight]{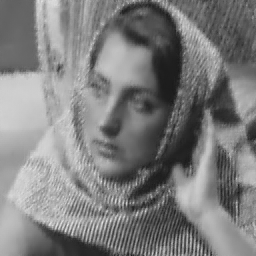}
\caption{\small Barbara reconstructions from a random 20\% of the pixels: corrupted image~(top), KFA~(center), BPFA~(bottom). KFA reconstructs the eyes and mouth better than BPFA. Also, the contrast in the stripe pattern is stronger in the KFA reconstruction. }
\label{fig:barbaraRecon}
\end{center}
\vskip -0.2in
\end{figure}

\subsection*{Dictionary Recovery}
\begin{figure}[!b]
\begin{center}
\centerline{\includegraphics[width=0.34\textheight]{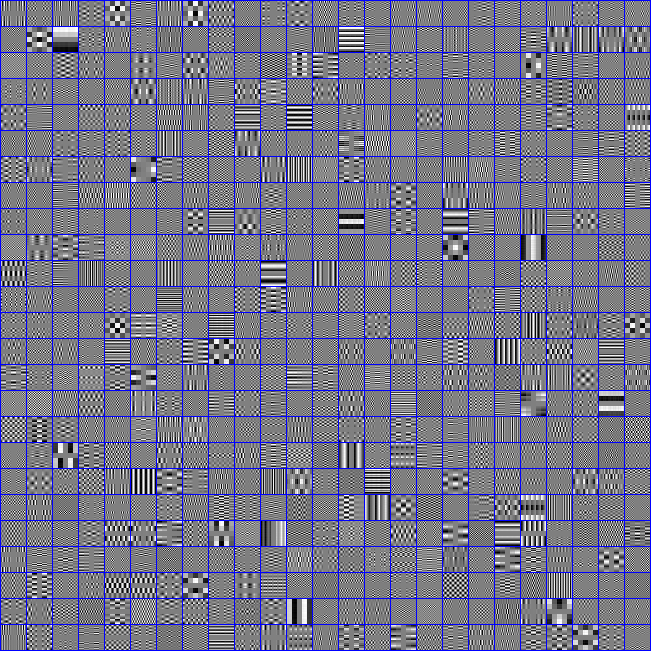}}\vskip 1mm
\centerline{\includegraphics[width=0.34\textheight]{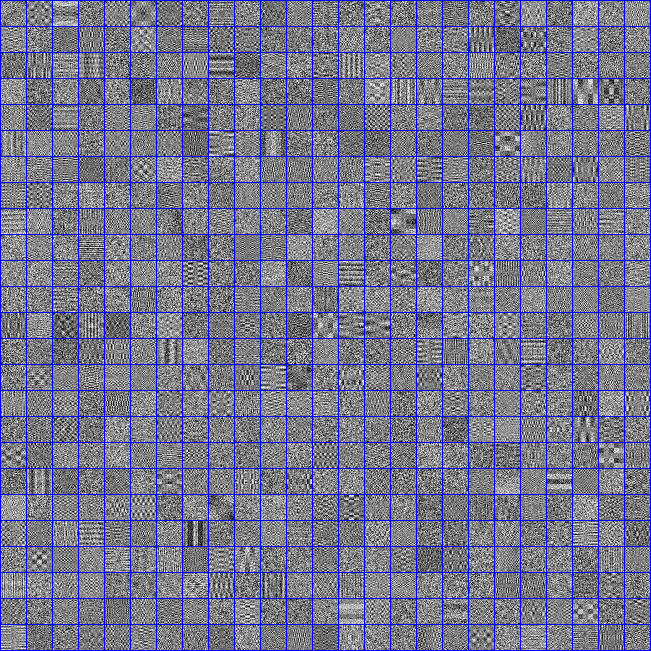}}
\caption{\small 4D-DCT example dictionaries collapsed to 2D: KFA~(top), BPFA~(bottom). Zoom-in for fine structure.}
\label{fig:dct}
\end{center}
\vskip -0.2in
\end{figure}

For this example, we assume the dictionary is composed of the $5\times5\times5\times5$ DCT filters. Note that each filter is rank-1 (separable). We generate 500~data samples using normal random weights for $\bm{S}$, enforce that only 8~nonzero elements are present in each column of $\bm{Z}$ and set the noise variance to~0.01. We use the same settings for BPFA and KFA, with $R=1$. The dictionary size is $K=625$, the same as the number of DCT filters. We ran both models for 100~iterations. Figure~\ref{fig:dct} shows the learned dictionaries from each algorithm. KFA learns the DCT filters. We find that even when we set $R$ to the expected rank $R_E=30$, the KFA dictionary atoms are low-rank. The BPFA dictionary looks noisy (high-rank structures) while both of the KFA dictionaries have low-rank atoms.
The RMSE for the three models is given in table \ref{tab:dct}. We can see from the dictionary and the RMSE that KFA is discovering the multi-way structure of the data. The number of samples is smaller than the dictionary, yet KFA is still able to find a well structured dictionary and achieve a lower error.
\begin{table}[!h]
\vskip -0.1in
\begin{center}
\caption{\small RMSE on 4D-DCT example} 
\label{tab:dct}
\vskip 2mm 
\begin{small}
\begin{sc}
  \begin{tabular}{l| c | c | c }
      & BPFA    & KFA $R=1$&  KFA $R=30$\\\hline
  RMSE& 0.03795&  0.01387& 0.01095 \\
  \end{tabular}
\end{sc}
\end{small}
\end{center}
\vskip -0.1in
\end{table}

\end{document}